\documentclass[preprint,12pt]{elsarticle}

\usepackage{amssymb}
\usepackage{amsmath}
\usepackage{multirow}
\usepackage{multicol}
\usepackage{bbding}
\usepackage[colorlinks,linkcolor=red]{hyperref}

\journal{Neurocomputing}

\begin{document}

\begin{frontmatter}



\title{Unsupervised Dense Nuclei Detection and Segmentation with Prior Self-activation Map For Histology Images}

\author[1,2]{Pingyi Chen}
\author[2]{Chenglu Zhu}
\author[2]{Zhongyi Shui}
\author[2]{Jiatong Cai}
\author[2]{Sunyi Zheng}
\author[2]{\\Shichuan Zhang}
\author[2,5]{Lin Yang}

\address[1]{{School of Computer Science and Technology, Zhejiang University},
            {Hangzhou},
            {China}}

\address[2]{{School of Engineering, Westlake University},
            {Hangzhou},
            {China}}
            
\fntext[4]{This work was supported in part by China Postdoctoral Science Foundation (2021M702922).}

\fntext[5]{Corresponding author}

\fntext[6]{Email address: chenpingyi@westlake.edu.cn (Pingyi Chen), yanglin@westlake.edu.cn (Lin Yang)}

\begin{abstract}
The success of supervised deep learning models in medical image segmentation relies on detailed annotations.
However, labor-intensive manual labeling is costly and inefficient, especially in dense object segmentation. 
To this end, we propose a self-supervised learning based approach with a Prior Self-activation Module (PSM)  that generates self-activation maps from the input images to avoid labeling costs and further produce pseudo masks for the downstream task.
To be specific, we firstly train a neural network using self-supervised learning and utilize the gradient information in the shallow layers of the network to generate self-activation maps. Afterwards, a semantic-guided generator is then introduced as a pipeline to transform visual representations from PSM to pixel-level semantic pseudo masks for downstream tasks. 
Furthermore, a two-stage training module, consisting of a nuclei detection network and a nuclei segmentation network, is adopted to achieve the final segmentation. 
Experimental results show the effectiveness on two public pathological datasets. Compared with other fully-supervised and weakly-supervised methods, our method can achieve competitive performance without any manual annotations.
\end{abstract}




\end{frontmatter}


\section{Introduction}
\label{sec:introduction}
Nuclei recognition serves a key role in  exploiting the microscopy image for disease diagnosis. Clear and accurate nuclei shapes provide rich details: nucleus structure, nuclei counts, and nuclei density of distribution. 
Hence, pathologists are able to conduct a reliable diagnosis under the information from the segmented nuclei, which also improves their experience of routine pathology workflow
\cite{elston1991pathological,le1989prognostic}.

\begin{figure}
\centering
\includegraphics[width=\linewidth]{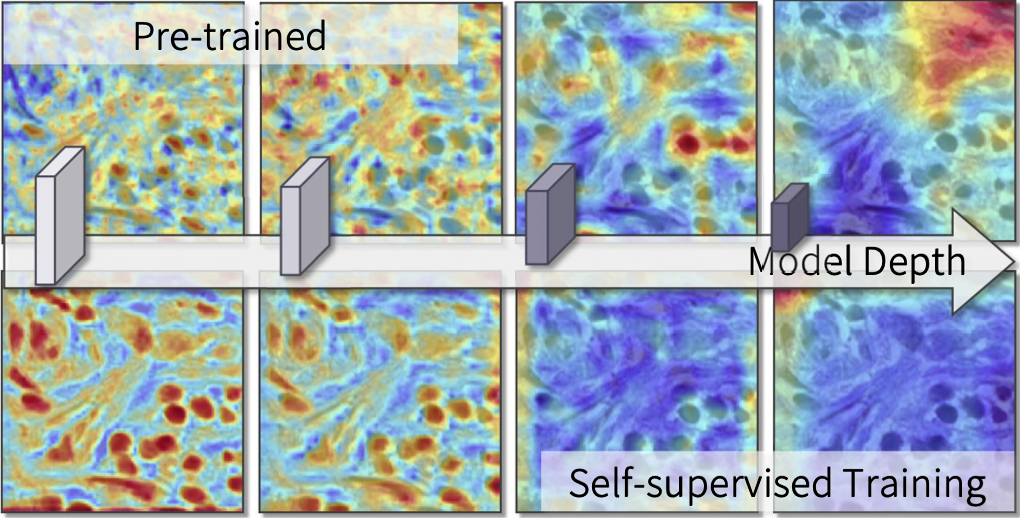}
\caption{The activation maps in different depths of the ImageNet pretrained network and self-supervised network. }
\label{fig1}
\end{figure}

In recent years, the advancement of deep learning has facilitated significant success in medical image segmentation 
\cite{naylor2018segmentation,liu2019nuclei,2015unet}.
In spite of the excellent results provided by these methods, extensive manual annotation is indispensable to obtain the excellent result. 
The fully-supervised property requires massive manual labels.
Especially when labeling histopathology images, a large number of nuclei are required to be labeled, which results in inefficient annotating processes. Furthermore, due to  the privacy of this type of data, and the variability of reading experience among pathologists, the limited annotations affect the model's effectiveness in real scenarios.

To alleviate this problem, work has been devoted to reducing dependency on the manual annotations in the analysis of histopathology images recently \cite{2020Weakly,qu2019weakly}. One of the common ideas to solve this problem is to decrease the completeness of annotations, i.e. weakly supervised learning. 

In terms of weakly-supervised segmentation in natural images, scribbles, bounding boxes, or classification labels are used to train the model. A popular line of work is the weakly supervised semantic segmentation based on the class activation map (CAM) \cite{2016Learning,gradcam}.
Recent studies \cite{2019FickleNet,2020Self,2020Attention}  aim to generate refined segmentation masks through coarse segmentation results, which are obtained by extracting highlighted regions generated by CAM. With lacking the more specific description of annotations, the map obtained tends to be more sensitive to the most discriminative region of the image. \cite{zhang2021complementary}. 
Therefore, the results of CAM are normally conducive for the detection of specific instances but may not be appropriate for the segmentation of densely distributed objects in microscopy images. To eschew this problem, we propose the prior self-activation map that can focus on the cell-level instances and need no annotations at the same time. 

With respect to the weakly supervised methods in histopathology images, widely-adopted ones are based on the point annotations which indicate the center of each nucleus \cite{qu2019weakly,2020Weakly}. Although such instance-level labels can be considered weak labels for cell segmentation and have achieved promising performance, using such labels still suffers from inaccurate boundaries. Moreover, annotating cell-wise labels is still a labor-intensive task due to the large number of objects required to be annotated. 

Regard to unsupervised nuclei recognition, traditional methods can segment the nuclei by clustering or morphological processing. But these methods suffer from worse performance than deep learning methods.
Some deep learning-based works utilize domain adaptation to realize unsupervised instance segmentation in biomedical images \cite{2019ddmrl,sifa,cycada}, which transfer the source domain containing annotations to the unlabeled target domain because of the similarity of both domains. 
However, it is not easy to obtain an annotated source dataset with little distribution difference for any target dataset.

Self-supervised learning  enables neural networks to extract  features without manual annotations, such as SimCLR \cite{chen2020simple} and MoCo \cite{he2020momentum}). Xie et al. \cite{instanceaware} also propose to use self-supervised learning to loose the need for annotations. These methods can obtain reliable representations by pre-training on a proxy task with unlabeled data. But when pre-trained models are transferred to downstream tasks, labels are still needed for finetuning. In our work, we develop a prior self-activation module to generate pseudo nuclei masks without any labels.

Our pipeline for label-free dense object segmentation can be broken down into the upstream task for learning semantic representations and the downstream task for the detection and segmentation of intensive visual cells. 
Apparently, it is a challenge to implement a downstream model for this advanced visual task without any annotations.
The key factor for the implementation of such a model is to observe model interpretations from the self-activation maps of the self-supervised pre-trained network, as shown in Fig. \ref{fig1}. Although the model generates deep semantic representations which are not sensitive to the boundaries of nuclei, its low-level semantic features from shallow layers are conducive to dealing with the downstream task. Therefore, this discovered correlation can establish a self-constraint path between upstream and downstream tasks, which is useful for dense object segmentation in self-supervised learning with unlabeled data.

Based on the above considerations, we proposed the Prior Self-activation Module (PSM) in this paper to realize the unsupervised nuclei segmentation.
Specifically, a network is initially trained under the supervision of prior labels such as rotation angles, instance-level contrastiveness, and average pixel values. 
Gradient information accumulated in the first layer of the network is then calculated and presented in the form of heat-maps. 
Secondly, high-quality pseudo masks used as supervision are generated by the semantic-guided generator (SGG), which fuses the interpretations of low-level features with original image information.
In addition, a two-stage training module is proposed to finetune the segmentation results.
In particular, the pixel-level pseudo masks from PSM serve as the supervision for the nuclei detection network (NDN). 
NDN generates a peak-shape probability map centered at each nucleus and the center points can be detected by searching the local extremum in the probability map. Based on the point detection results, the instance-level Voronoi labels and pixel-level segmentation masks jointly supervise the nuclei segmentation network (NSN). In the inferring process, NSN is directly applied for segmentation. In addition, the detection result can be obtained by post-processing the output of NDN.
Our main contributions are the following (the code is available at: \url{https://github.com/cpystan/Prior-Self-activation-Map}):

\begin{itemize}

    \item A novel pipeline for unsupervised nuclei detection and segmentation. 
    \item The Prior Self-activation Module (PSM) is proposed for use in unlabeled biomedical image data, which is able to generate self-activation maps for downstream segmentation tasks. Also, PSM has the potential in processing various types of medical images.

    \item The semantic-guided generator 
    is constructed, which  
    utilizes the integration of self-activation maps and original image information to generate pixel-level pseudo masks.
  
    \item The architecture shows competitive performance compared to other methods with full and weak supervision, which reveals the superiority of our proposed method.

\end{itemize}

\begin{figure*}
\includegraphics[width=\textwidth]{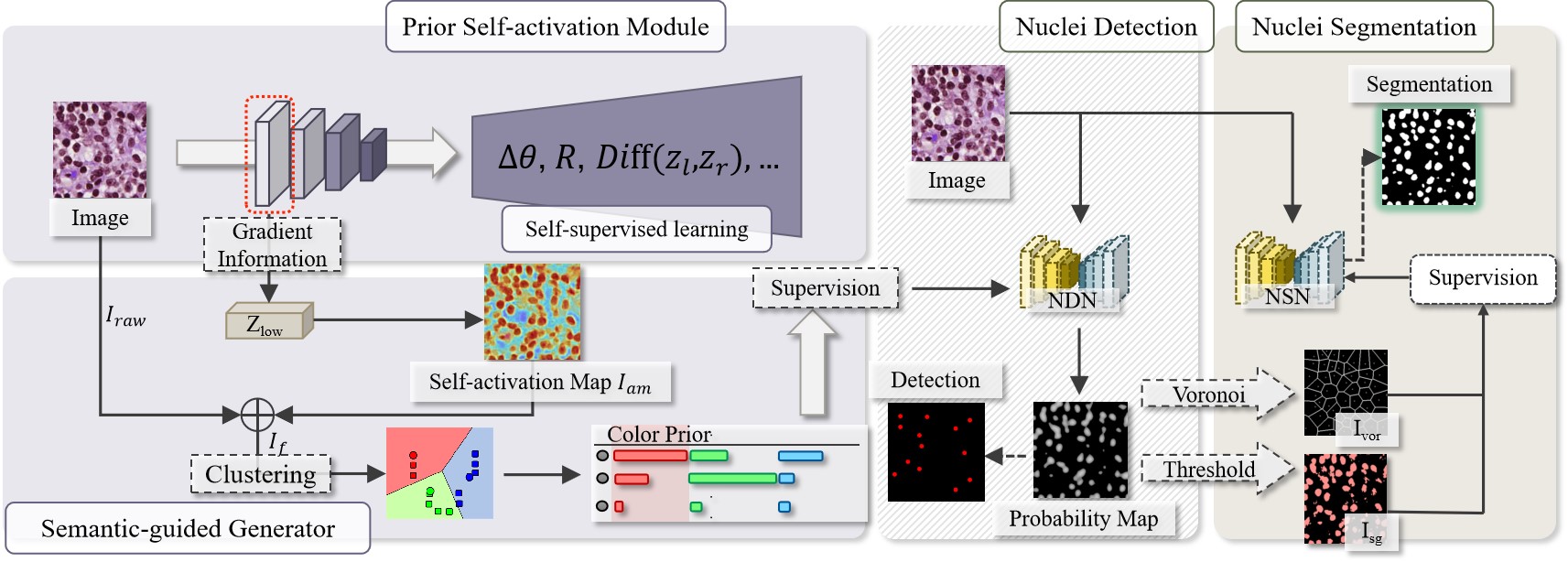}
\caption{The framework of our proposed model. Top left corner shows the prior self-activation module which minimizes the high-dimensional vectors from the input and its augmented view and uses low-level features in the network to generate self-activation maps. Bottom left presents the work process of the semantic-guided generator. The input and its low-level representations are fused and clustered followed by label reassignment with color prior information
. The right part is the nuclei detection and segmentation module which employ both pixel-level and object-level labels to train the backbone.  } \label{fig2} 
\end{figure*}

\section{Related Work}

In this section, we introduce the related works in two aspects:  (1) nuclei detection and (2) nuclei segmentation. 

\subsection{Nuclei detection}
Nuclei detection can provide much valuable clinical clues like  the numbers of each category and the distribution of cells.  Recent year have witnessed a rapidly growing performance based on deep learning methods. Convolution neural network shows its superiority over learning robust representations automatically without exploiting handcrafted visual features.

We group these nuclei detection methods into two lines. The first is those works using bounding box as annotations. Some methods designed for natural images which have shown great performance can be transferred to biomedical images with the bounding boxes. They train a classifier to make predictions for the patches that are selected with a sliding-window. Some early works have obtained promising performance for cell detection \cite{cirecsan2013mitosis,li2018deepmitosis,locality}. However, these bounding box-based approaches still have some drawbacks. The shape and scale variation give much trouble to designing suitable boxes for the cells. The other line of works is to detect nuclei with the aid of point annotations \cite{zhou2018sfcn,2020BCData}. Without predicting the bounding boxs surrounding cells, they learn to localize the nuclei centers. Point annotations can largely reduce the workload for annotating and provide sufficient clinical clues at the same time. Several works make predictions based on the density maps. These regression models outputs a probability map where the local extremums are 
selected as the predicted cell centers. However, their success relies on the quality of the density maps.  In our method, we detect the centers of the nuclei with the pseudo masks which are generated automatically.
 
\subsection{Nuclei segmentation} 
Deep convolutional neural networks have been applied in nuclei segmentation and obtained promising results.  Xing et al. \cite{xing} performed an iterative region merging approach for shape initialization and a local repulsive deformable model for robust nuclei segmentation. Kumar et al. \cite{2017monu} introduced a large public dataset and proposed a segmentation technique that lays an emphasis on the nuclear boundary identification, especially on those hard examples with overlapping.

In terms of nuclei segmentation with coarse labels, Qu et al. \cite{qu2019weakly} firstly proposed 
to apply instance-level annotations to generate pseudo masks for segmentation training. With only partial points, they first detect the complete point annotations with an iterative self-training procedure. Other works employed Sobel filters to generate pseudo edge labels for training \cite{yoo2019pseudoedgenet,2020Weakly} or trained a model to predict the probability map around the nuclei centers \cite{chamanzar2020weakly}. 
Those methods still have problems in dealing with crowded nuclei because of the lack of expressive instance-aware representations. 
Hou et al. \cite{hou2019robust} first proposed to train a nuclei segmentation model under the synthesis framework. 
Liu et al. \cite{liu2020pdam} aligned the cross-domain features and employed representation learning to facilitate instance-level feature adaptation. 
But their model performance suffers from the domain shift when a huge variance exists between the source domain and the target domain. 

\section{Framework}

The structure of our proposed method is demonstrated in Fig. \ref{fig2}. 
Firstly, the Prior Self-activation Module (PSM) is utilized to provide pseudo masks for subsequent tasks.
Specifically, the gradient of the network is exploited to generate the self-activation maps.
Next, a special semantic-guided generator (SGG) is proposed to transform self-activation maps to pixel-level pseudo masks.
Finally, a two-step training module, which contains a nuclei detection network (NDN) and a nuclei segmentation network (NSN), is introduced to achieve dense nuclei detection and segmentation.

\subsection{Proxy Tasks for Self-activation Map}
The main idea behind PSM is that CNNs with inductive biases have priority over some local features with certain properties in the image \cite{inductive}, especially for the nuclei with dense distribution and semi-regular shape.  
As shown in Fig. \ref{fig:layers}, the visualization result shows that shallow layers in the self-trained network are suitable for nuclei detection. 
Due to the relatively small receptive field, the  maps of shallow features in the self-trained network are more conducive to local descriptions. 

Although low-level information in the self-trained network is relatively stable, the availability of self-activation maps is highly relevant to the proxy task in self-supervised learning. Here, we have experimented with several basic proxy tasks below, as shown in Fig. \ref{fig:layers}:

\subsubsection{ImageNet pre-training} It is not strictly a self-supervised strategy. It is  straightforward to exploit the models pre-trained on natural images. In this strategy, we first train our model on ImageNet and then feed nulcei images to generate the self-activation map. 

\subsubsection{Mean pixel value} For each input image, the average pixel value $R$ is calculated first and used as the training target. Thus, the proxy task here is to predict the mean value of the input image. However, during the experiment, we found that the model can not converge on this proxy task well. Since that, it is not discussed in the following paragraph.

\subsubsection{Rotation angle} For each sample, before being fed into the model, will be randomly rotated by 0, 90, 180, or 270 degrees. Therefore, the proxy task is to predict the rotated angle $\Delta \theta$ of the input. 

\subsubsection{Contrastiveness} Following the contrastive learning \cite{chen2020simple} methods, the network is encouraged to distinguish between different patches. For each image, its augmented view will be regarded as the positive sample, and the other image sampled in the training set is defined as the negative sample. The network is trained to minimize the distance between positive samples and maximize the one between the negative sample and the input image.

\subsubsection{Similarity} LeCun et al. \cite{siamese} proposed a Siamese network to train the model with a similarity metric. Here, we also adopted a weight-shared network to learn the similarity discrimination task. In specific, the pair of samples (each input and its augmented view) will be fed to the network, and then embedded as high-dimensional vectors $Z_l$ and $Z_r$ in the high-dimensional space, respectively. 
Based on the similarity measure, $L_{dis}$ is introduced to reduce the distance between them for extracting common features, which is denoted as,
\begin{equation}
    L_{dis}(Z_l,Z_r)=abs(Z_l-Z_r).
\end{equation}

In our scenario, we discovered that similarity discrimination based self-supervision generates the best self-activation maps among the aforementioned strategies. Therefore, all of our following nuclei detection and segmentation experiments are based on this proxy task.

\subsection{Self-activation Map Generation}
An intuitive way to obtain the pixel-level semantic masks for supervising downstream tasks is the utilization of model interpretation, 
such as class activation mapping (CAM), which can highlight the region of interest.
Some previous works \cite{2019FickleNet,2020Self,2020Attention} realize the weakly-supervised segmentation by utilizing the visual presentation of CAM,
which is generated from gradient information of the last convolutional layer. 
Nevertheless, CAM-based approaches can not be directly applied in the unsupervised nuclei detection or segmentation due to the two following problems:

\begin{itemize}
    \item CAM and CAM variants still need manual annotations for supervision like image classification labels, etc. 
    \item Activation maps produced by CAMs can not provide accurate boundaries for dense objects.
\end{itemize}

To deal with the first problem, we have introduced self-supervised learning to train the network without annotations. 
In terms of the second problem, we propose to exploit the gradient information in the first convolutional layer. To distinguish from the existing activation maps, we name it \textit{prior self-activation map}.
The following experiment (see Fig. \ref{fig:layers}) also reveals that low-level features for different proxy tasks are suitable for pixel-level tasks, especially for the dense object segmentation which depends on the regional description. Therefore, compared to CAMs, our prior self-activation map (PSM) has two advantages: 1) no labels, and 2) features from the shallow layer. 

In specific, the self-supervised model is constructed by sequential blocks which contain several convolutional layers, batch normalization layers and activation layers. 
The self-activation map of a certain block can be obtained by nonlinearly mapping the weighted feature maps $A^k$ :
\begin{equation}
I_{am}= ReLU(\sum_k {\alpha}_k A^k),
\end{equation}
where $I_{am}$ is the self-activation map. $A^k$ indicates the $k$-th feature map in the selected layer. ${\alpha}_k$ is the weight of each feature map, 
which is defined by global-average-pooling the gradients of output $z$ with regard to $A^k$:
\begin{equation}
    {\alpha}_k = \frac{1}{N} \sum_i \sum_j \frac{\partial {z}}{\partial {A_{ij}^k}},
\end{equation}
where $i$,$j$ denote the height and width of output, respectively, and $N$ indicates the input size.

\subsection{Semantic-guided Generator}
 We construct a semantic-guided generator (SGG) which converts self-activation maps to pseudo masks. 
In SGG, the original information is added to the self-activation maps to strengthen the detailed features.
It is defined as:
\begin{equation}
    {I_{f}}= I_{am} + \beta \cdot I_{raw},
\label{detail}
\end{equation}
where ${I_{f}}$ denotes the fused semantic map, $I_{raw}$ is the raw image,  $\beta$ is the weight of $I_{raw}$. 

Considering the color prior, an unsupervised clustering method K-Means is selected to directly split all pixels into several clusters and obtain foreground nuclei and background pixels. 
Given the self-activation map $I_f$ and its $N$ pixel features $F=\{f_i | i\in \{1,2,...N\}\}$, $N$ features are partitioned into $K$ clusters $S = \{S_i | i \in \{1,2,...K\}  \}$, $K<N$. The goal is to find $S $ to reach the minimization of within-class variances as follows:
\begin{equation}
    min \sum_{i=1}^K \sum_{f_j\in S_i} ||f_j - c_i||^2,
\end{equation}
where $c_i$ denotes the centeroid of each cluster $S_i$. 

Nevertheless, due to the randomness of K-Means with setting the initial seeds, these clusters often suffer from the problem of label inconsistency, which means the cluster labels for certain instances are not always the same.
Hence, it is unavoidable to reassign label for target cluster. Apparently, the prior knowledge of color corresponds with salient regions of the self-activation map. In other words, foreground pixels tend to have a high value in red channel because they are highlighted by the map $I_{am}$.
Such color prior brought out by the property of heat-map is universal in our unsupervised framework, which means we do not need to resort to other priors in clustering if the dataset changes. 

\subsection{Nuclei detection and segmentation} 

In this section, we elaborate on nuclei detection and segmentation with the aid of generated pseudo masks. The process comprises two modules: Nuclei Detection Network (NDN) and Nuclei Segmentation Network (NSN). NDN is applied  to produce the probability map under the supervision of pseudo masks. By post-processing the map, the detection result can be obtained.
NSN is trained as the final inference model using the pixel-wise and instance-wise supervision from previous modules. 

\subsubsection{Nuclei Detection}
In the step of nuclei detection, a detection network is set as the probability map generator, which is trained under the supervision of pseudo masks from SGG. 
Subsequently, labels of two levels of supervision can be obtained from the generated probability maps.

The pixel-level binary labels $I_{sg}$ focus on foreground and background pixels with high confidence. 
Foreground labels are assigned to the pixels whose probability value is above the higher threshold $T_{fg}$ and the pixels below the lower threshold $T_{bg}$ are background, which can be formulated as below:

\begin{equation}
I_{sg} =\left\{
\begin{array}{l}
1, \quad p_{(i,j)}>T_{fg},\\
0, \quad p_{(i,j)}<T_{bg},\\
-1, \quad otherwise.\\
\end{array}
\right.
\end{equation}

Due to the lack of instance-level supervision, the model does not perform well in distinguishment between adjacent objects prone to connect with each other.
To further reduce errors and uncertainties, the Voronoi map is constructed to provide complementary instance-wise information.

The nuclei centroid can be treated as the seed point in the Voronoi diagram. 
Depending on the dense distribution and semi-regular shape of the nuclei, the Voronoi edge can accurately separate close nuclei. In training, the edges are labeled as background and the seed points are denoted as foreground. Other pixels are ignored during training. 

The seed points for Voronoi maps can be obtained by searching local extremums in the probability map, which is described below:
\begin{equation}
T_{(m,n)}=\left\{
\begin{array}{l}
1, \quad p_{(m,n)}>p_{(i,j)},\quad {\forall}(i,j)\in D_{(m , n)},\\
0, \quad otherwise,
\end{array}
\right.
\end{equation}
where $T_{(m,n)}$ denotes the predicted center label at location of $(m,n)$, $p$ is the value of the probability map and $D_{(m , n)}$ indicates the neighborhood of point $(m,n)$. Then we can obtain the Voronoi map $I_{sg}$. It is worth noting that $T_{(m,n)}$ is exactly the detection result.

\subsubsection{Nuclei Segmentation}
In this step, we train the segmentation model NSN with the above two types of labels. 
The mixture between pixel-level features and instance-level features facilitates the overall performance. 
The training loss function can be formulated as below,
\begin{equation}
\begin{split}
L=  \lambda [ y \log I_{vor} &+ (1-y) \log (1-I_{vor})] \\
                                  &+ (1 - y)\log (1-I_{sg}),
                                  \end{split}
\label{vor}
\end{equation}
where $\lambda$ is the partition enhancement coefficient. In our experiment, we discovered that false positives  hamper the effectiveness on instance segmentation due to the ambiguity of nuclei boundaries. Since that, only the background of $I_{sg}$ will be concerned to eliminate the influence of false positives in instance identification.

For inference, NSN can be directly used to realize the end-to-end nuclei segmentation. 

\section{Experiments}

\subsection{Dataset and Metric}
\subsubsection{Dataset} We validated the proposed method on the public dataset of Multi-Organ Nuclei Segmentation (MoNuSeg) \cite{2017monu} and Breast tumor Cell Dataset (BCData) \cite{2020BCData}.
MoNuSeg consists of 44 images of size 1000 $\times$ 1000 with around 29,000 nuclei boundary annotations. All of them are H${\&}$E stained tissue images from various organs captured under the magnification of 40X. These images are split into 30, 14 for training and testing.
BCData is a public large-scale breast tumor dataset containing 1338 immunohistochemically Ki-67 stained images of size 640 $\times$ 640. These images are split into 803, 133, and 402 for training, validation, and testing.
A total number of 181,074 tumor cells are labeled as point annotations at the center of each target cell.

\subsubsection{Evaluation Metrics}
In our experiments on MoNuSeg, F1-score and IOU are employed to evaluate the segmentation performance. Denote $TP$, $FP$, and $FN$ as the number of true positives, false positives, and false negatives. Then F1-score and IOU can be defined as: $F1=2TP/(2TP+FP+FN)$, $IOU=TP/(TP+FP+FN)$. In addition, common object-level indicators such as Dice coefficient and Aggregated Jaccard Index \cite{2017monu} are also considered to assess the segmentation performance. 

The object-level Dice coefficient represents the overlapped measure between the predicted object and the corresponding ground truth as follows,
\begin{equation}
   Dice_{obj}= \frac{1}{2} \left [ \sum_{i=1}^{n_{gt}}\gamma _{gt}(i)D_{gt}(i) + \sum_{j=1}^{n_{p}}\gamma _{p}(j)D_{p}(j)\right], \\
\end{equation}
 where ${D_{gt}(i)=Dice(g_{i}, p^{*} (g_{i}))}$ and ${D_{p}(j)=Dice(g^{*}(p_{j}) , p_{j}) }$.
${ p^{*} (g_{i})}$ indicates the prediction has the maximum overlapp ed area with ${g_{i}}$, conversely, ${g^{*}(p_{j})}$ denotes the the maximum overlapped region of ground truth to ${p_{j}}$.
Besides, ${\gamma _{gt}(i)}$ and ${\gamma _{p}(j)}$ is the proportion calculated for pixels in each object to the total pixels of all objects in ground truth and prediction, respectively.

The Aggregated Jaccard Index (AJI) takes unmatched instances into account and is defined as follows,
\begin{equation}
AJI=\frac{\sum_{i=1}^{n_{gt}}\left | g_{i}\bigcap p(g_{i})  \right | }{\sum_{i=1}^{n_{gt}}\left | g_{i}\bigcup  p(g_{i})  \right | + {\sum_{j}^{n_{um} }\left | p_{j} \right |  } },
\end{equation}
where ${n_{um}}$ represents predictions that are not matched with any ground truth objects. $n_{gt}$ indicates the total number of ground truth nuclei.

In the experiment on BCData, precision (P), recall (R), and F1-score are used to evaluate the detection performance. Predicted points will be matched to  ground-truth points one by one. And those unmatched points are regarded as false positives. Precision and recall can be defined as: $P=TP/(TP+FP)$, and $R=TP/(TP+FN)$. In addition, we introduce MP to evaluate the cell counting results. 'MP' denote the mean average error of positive cell numbers and '$\downarrow$' means that the lower value shows the better performance.

\begin{table}[th] 
  \centering
  \caption{Results on MoNuSeg. According to requirements of training, various levels of supervision are selected from localization (Loc) and contour (Cnt). * indicates model is trained from scratch with the same hyperparameter as ours.}
  \resizebox{\linewidth}{!}
{
  \begin{tabular}{l|cc|cc|cc}
   \hline
  \multirow{2}{*}{Methods} & \multirow{2}{*}{Loc} & \multirow{2}{*}{Cnt} & \multicolumn{2}{c|}{Pixel-level} & \multicolumn{2}{c}{Object-level} \\
  \cline{4-5} \cline{6-7}
  ~ & ~ & ~ & IoU & F1 score & Dice & AJI \\
  \hline
    ResUnet*\cite{2017resunet} & \CheckmarkBold & \CheckmarkBold & 0.599 & 0.735 & 0.709 & 0.504 \\
    DeepLab v3+*\cite{2018Encoder} & \CheckmarkBold & \CheckmarkBold & 0.560 & 0.742 & 0.682 & 0.458 \\
    Unet*\cite{2015unet} & \CheckmarkBold & \CheckmarkBold & 0.606 & 0.745 & {0.715} & 0.511 \\
    MedT\cite{2021medt} & \CheckmarkBold & \CheckmarkBold & {0.662} & {0.795} & - & - \\
    Hover-Net\cite{graham2019hover} & \CheckmarkBold & \CheckmarkBold & - & - & 0.826 & 0.618 \\
    \hline
    Yoo et al.\cite{yoo2019pseudoedgenet}& \CheckmarkBold & \XSolidBrush & 0.614 & - & - & - \\
    Qu et al.\cite{qu2019weakly}& \CheckmarkBold & \XSolidBrush & 0.579 & 0.732 & 0.702 & 0.496 \\
    Tian et al.\cite{2020Weakly}& \CheckmarkBold & \XSolidBrush & 0.624 & 0.764 & 0.713 & 0.493 \\
    \hline
        CellProfiler \cite{cellprofiler} & \XSolidBrush & \XSolidBrush & - & 0.404 & 0.597 & 0.123 \\
            Fiji\cite{fiji} & \XSolidBrush & \XSolidBrush & - & 0.665 & 0.649 & 0.273 \\
            
    DDMRL\cite{2019ddmrl} & \XSolidBrush & \XSolidBrush & - & 0.700 & - & 0.464 \\
    SIFA\cite{sifa} & \XSolidBrush & \XSolidBrush & - & 0.699 & - & 0.466 \\
    CyCADA\cite{cycada} & \XSolidBrush & \XSolidBrush & - & 0.705 & - & 0.472 \\
    \hline
    \textbf{Ours} & \XSolidBrush & \XSolidBrush & 0.610 & 0.750 & 0.724 & {0.532} \\
  \hline
  \end{tabular}}
   

\label{tab:comp}
\end{table}

\begin{table}[th]
  \centering
  \caption{Results on BCData. According to requirements of training, various levels of supervision are selected from localization (Loc) and contour (Cnt).}
\resizebox{\linewidth}{!}
{
 \begin{tabular}{l | cc | c | ccc|c}
  \hline
  Methods & Loc & Num & Backbone & P & R & F1 score&MP$\downarrow$\\
  \hline
    CSRNet\cite{csrnet} & \CheckmarkBold & \CheckmarkBold & ResNet50 &0.824 & 0.834& 0.829  &9.24 \\
    SC-CNN\cite{snpcnn} & \CheckmarkBold & \CheckmarkBold & ResNet50 &0.770 & 0.828& 0.798   & 9.18 \\
    U-CSRNet\cite{2020BCData} & \CheckmarkBold & \CheckmarkBold & ResNet50 & 0.869  & 0.857    &   0.863     & 10.04 \\
    TransCrowd\cite{transcrowd}& \XSolidBrush & \CheckmarkBold & Swin-Transformer&-&-&- & 13.08 \\
    \hline
    \textbf{Ours} & \XSolidBrush & \XSolidBrush & ResUnet& 0.855& 0.771 & 0.811& 8.89 \\
  \hline
  \end{tabular}}

\label{tab:comp1}
\end{table}

\begin{figure}
 \centering
\includegraphics[width=0.8\linewidth]{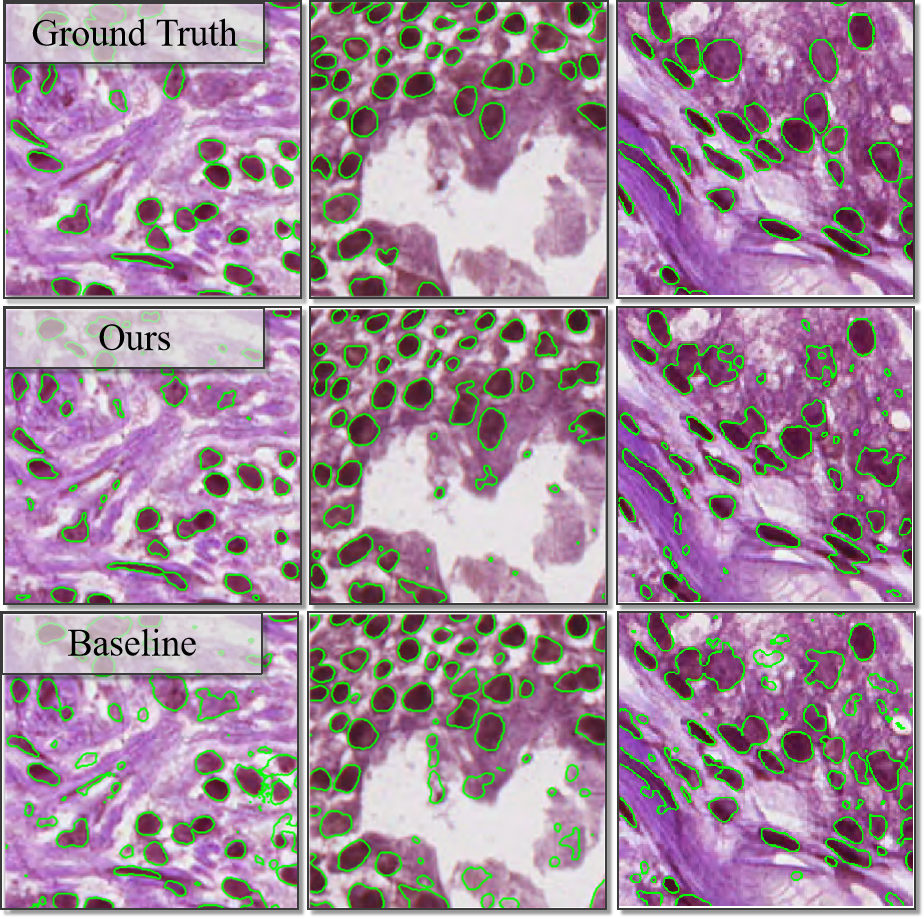}\\
 \caption{Visualization on MoNuSeg. The first row shows the ground truth. The second row presents the segmentation result of our method, and the final row is the output of ResUnet as the baseline.} \label{fig:cmp}
\end{figure} 

\subsection{Implementation Details}

Details about the structure of the proposed approach are described below. 

\subsubsection{Prior Self-activation Module}

Res2Net101 \cite{res2net} is adopted as the backbone of the PSM with random initialization of parameters, and the 'Similarity' discrimination is selected as the proxy task for self-supervised learning. Specifically, a paired input consists of the original image and the augmented view. And the distance between their high-dimensional features extracted by the PSM module is minimized by Adam optimizer with the initial learning rate of $1e^{-4}$ during $100$ epochs.

\subsubsection{Semantic-guided Generator}

In SGG, the fused semantic map is constructed by mixing heatmap and weighted raw image generated by PSM. The weights $\beta$ are set to  $2.5$ and $4$ for training in MoNuSeg and BCData, respectively. 
Pixels of the fused semantic map will be decoupled into three piles by K-Means.

\subsubsection{Nuclei Detection and Segmentation}
The following segmentation is constructed with two ResUnet models with ResNet-34 as the backbone. 
NDN is trained only under the supervision of the pixel-level semantic guidance with CrossEntropy loss. 
The threshold for background and foreground are all set to 0.6.
In addition, NSN is guided with the segmentation map and Voronoi map from the former prediction under the joint training loss. 
All models are trained by the Adam optimizer for $100$ epochs with the initial learning rate of $1e^{-4}$.

\begin{figure}
\centering
\includegraphics[width=\linewidth]{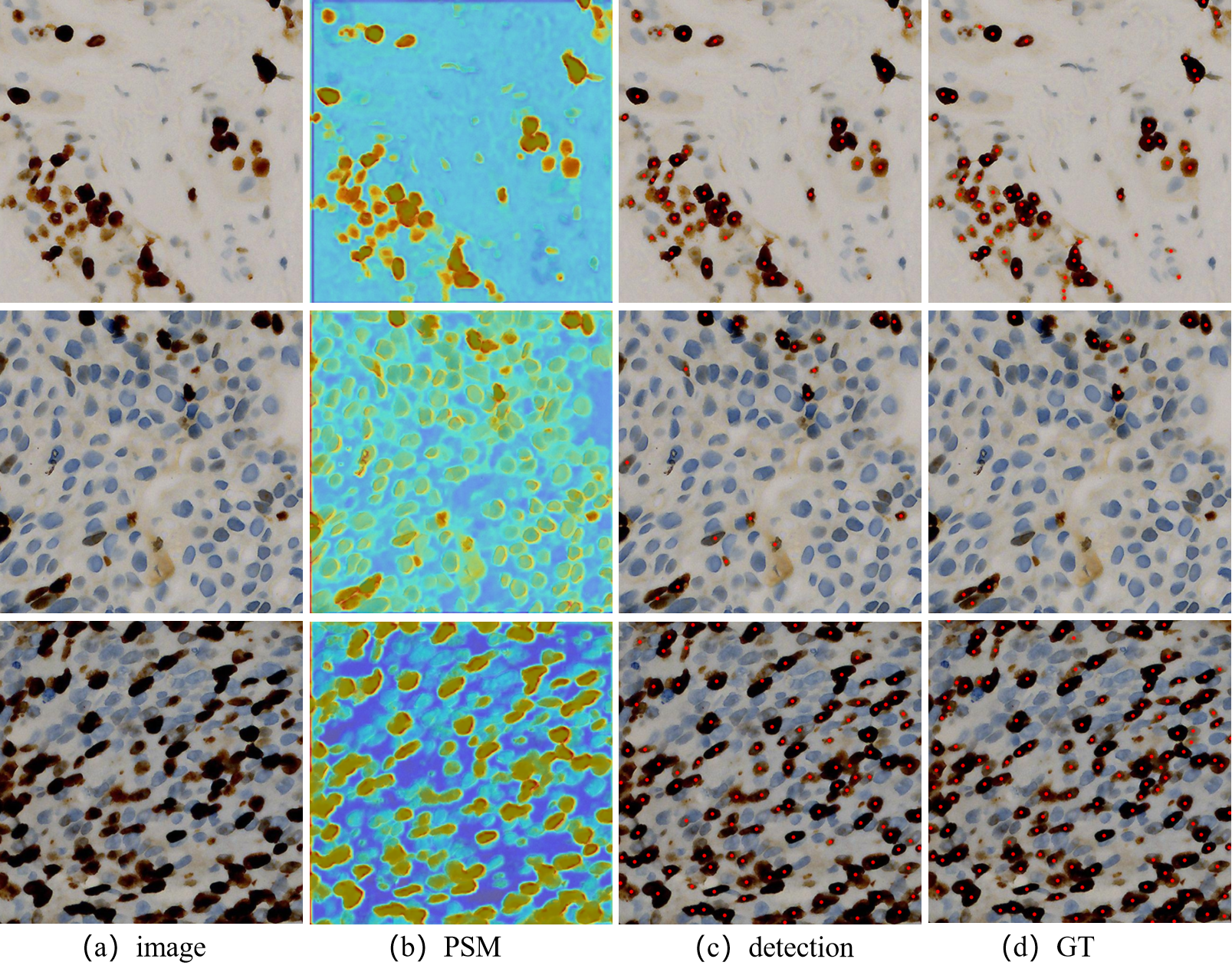}

\caption{Typical detection results on the BCD dataset. (a) example images, (b) prior self-activation maps of the images, (c) detection results on the images, and (d) ground truth. Points indicate the centroid of the nuclei. } \label{fig:bcd}
\end{figure}

\subsection{Results}
In MoNuSeg Dataset, five fully-supervised methods Unet \cite{2015unet}, ResUnet \cite{2017resunet},  DeepLab-V3+ \cite{2018Encoder}, 
MedT \cite{2021medt}, and Hover-Net \cite{graham2019hover} based on provided localization and contours are adopted to estimate the upper limit as shown in the first four rows of Table \ref{tab:comp}.
Three weakly-supervised models trained with point annotations are also performed to compare with ours.
Without the aid of labels, our method still produces competitive results except for MedT and Hover-Net. MedT is elaborately designed based on Transformer. Hover-Net introduces distance maps to separate clustered nuclei. Compared with the method \cite{2020Weakly} fully exploiting localization information, ours can achieve better performance without any annotations in object-level metrics (AJI).
In addition, two unsupervised methods using traditional image processing tools \cite{fiji,cellprofiler} and three unsupervised domain adaptation methods \cite{2019ddmrl,sifa,cycada} are compared.
It also shows our advantage in the instance-level identification as shown in Fig. \ref{fig:cmp}, adjacent objects which are easily confused can be separated successfully by our method.

Following the benchmark of BCData, metrics of detection and counting are adopted to evaluate the performance as shown in Table \ref{tab:comp1}.
The first three methods are the weakly supervised counting method based on the prediction of density maps generated with dot annotations.

Furthermore, TransCrowd \cite{transcrowd} with the backbone of Swin-Transformer is employed as the weaker supervision by utilizing directly the count number regression. 
By contrast, even without any annotation supervision, compared to CSRNet \cite{csrnet}, NP-CNN \cite{snpcnn} and U-CSRNet \cite{2020BCData}, our proposed method still achieved competitive performance. Especially in terms of MP, our model surpasses all the baselines.

\begin{figure}
    \centering
    \includegraphics[width=\linewidth]{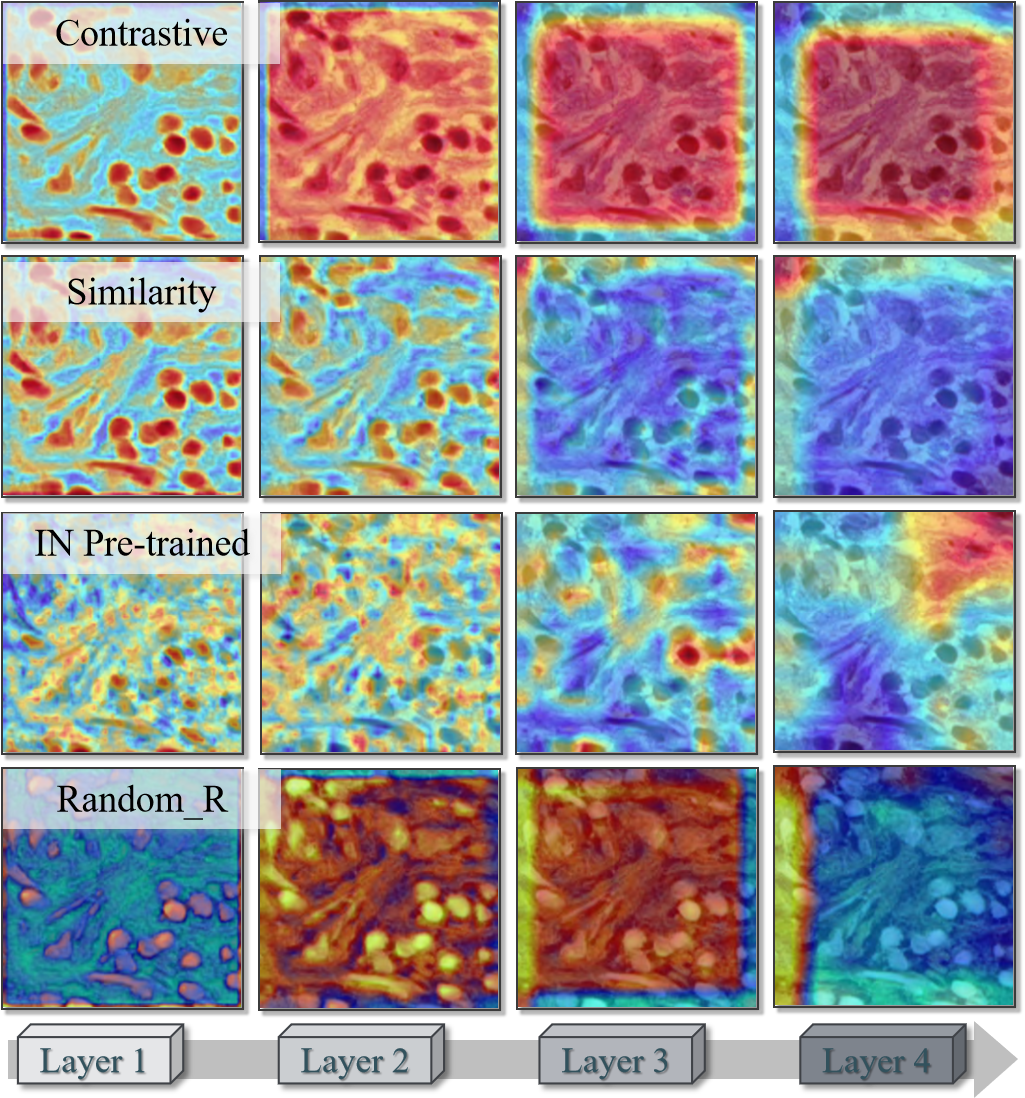}\\
    \caption{Comparison between self-activation maps in depths using different strategies. The self-activation map from the first layer trained with the 'Similarity' strategy shows the best visual features.  }
\centering
\label{fig:layers}
\end{figure}
\section{Analysis}

\subsection{Proxy Task} 

\begin{figure}
\centering
\includegraphics[width=\linewidth]{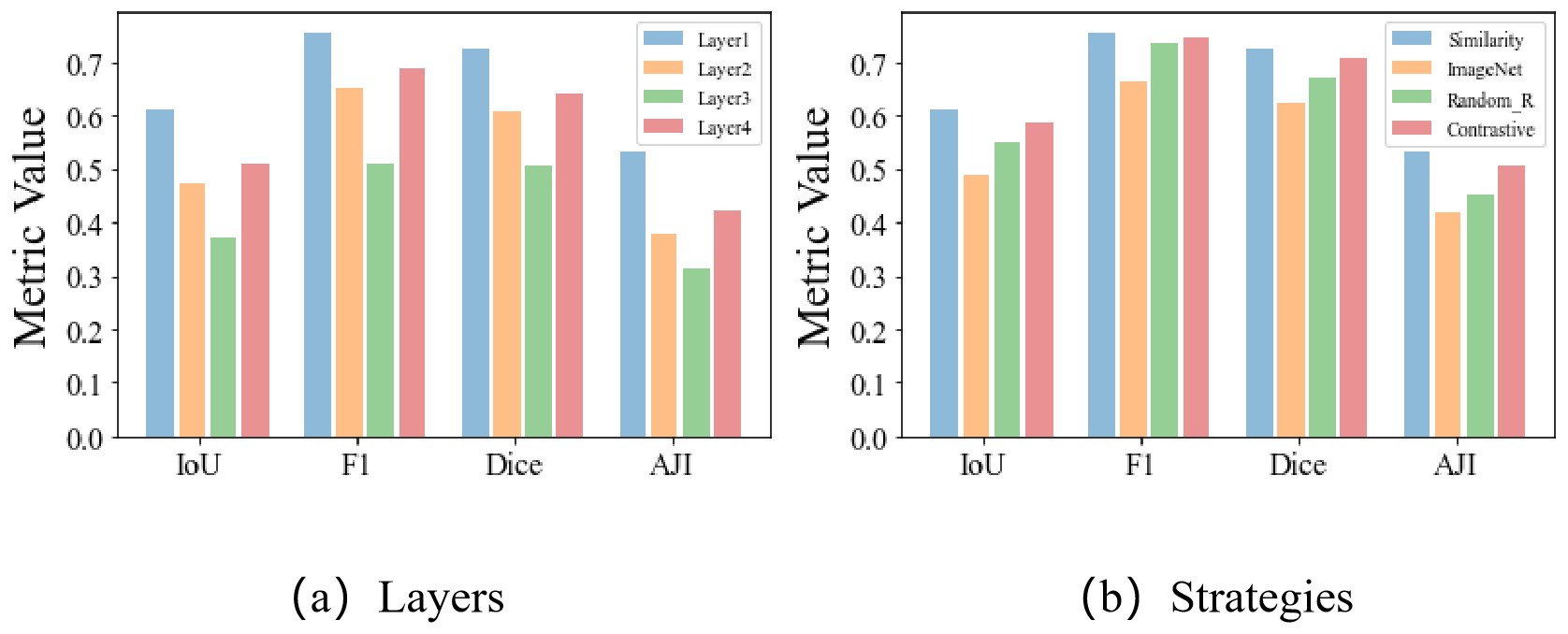}
\caption{Analysis of layer selection and self-supervised learning strategy selection in PSM.}
\label{fig:layers_proxy}
\end{figure}

We experimented with different types of proxy tasks as described in section II. 
All models are trained under the same settings, and their evaluation results are shown in Fig. \ref{fig:layers_proxy}.

Apparently, without the supervision of annotations, only relying on the pre-trained models with external data can not improve substantially the results of subsequent segmentation. 
From Fig. \ref{fig:layers}, the self-activation map generated from the ImageNet pre-trained network shows low positioning accuracy for dense objects. The large inter-domain variations exist and hamper the adaptation from natural images to medical images.
The 'Contrastiveness' strategy also fails to present a significant improvement. The high intra-domain similarity hinders the comparison between constructed image pairs. 
Unlike natural image datasets containing diverse samples, the minor inter-class differences in biomedical images may not exploit fully the superiority of contrastive learning whose success largely  depends on the variety between images.
Additionally, the model for the prediction of rotation angles performs the worst because such biomedical images are insensitive to rotation. Through experiments, we find that the strategy of 'Similarity' shows the best performance. 

\begin{figure}
\centering
\includegraphics[width=\linewidth]{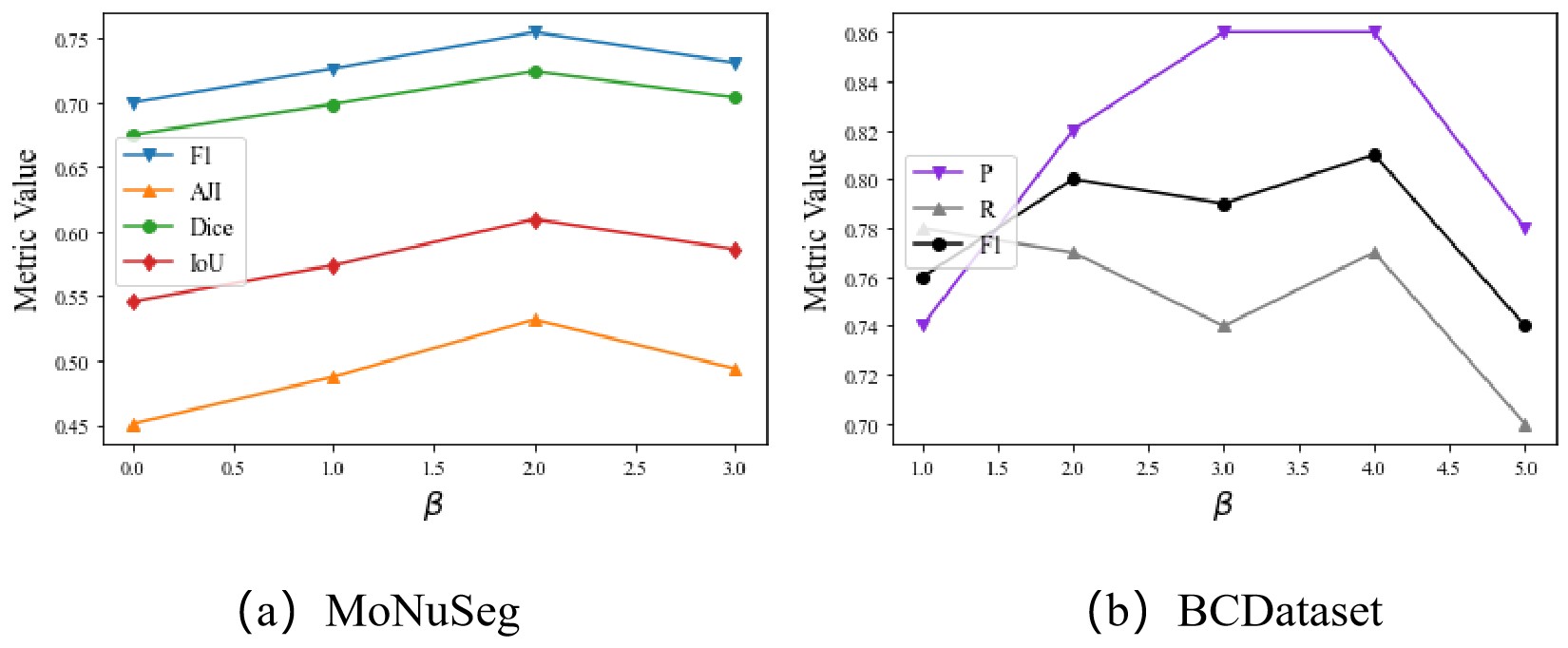}
\caption{The effect of hyperparameter $\beta$ on MoNuSeg and BCData datasets.  }
\label{fig:beta_metric}
\end{figure}

\begin{figure*}
\centering
\includegraphics[width=\textwidth]{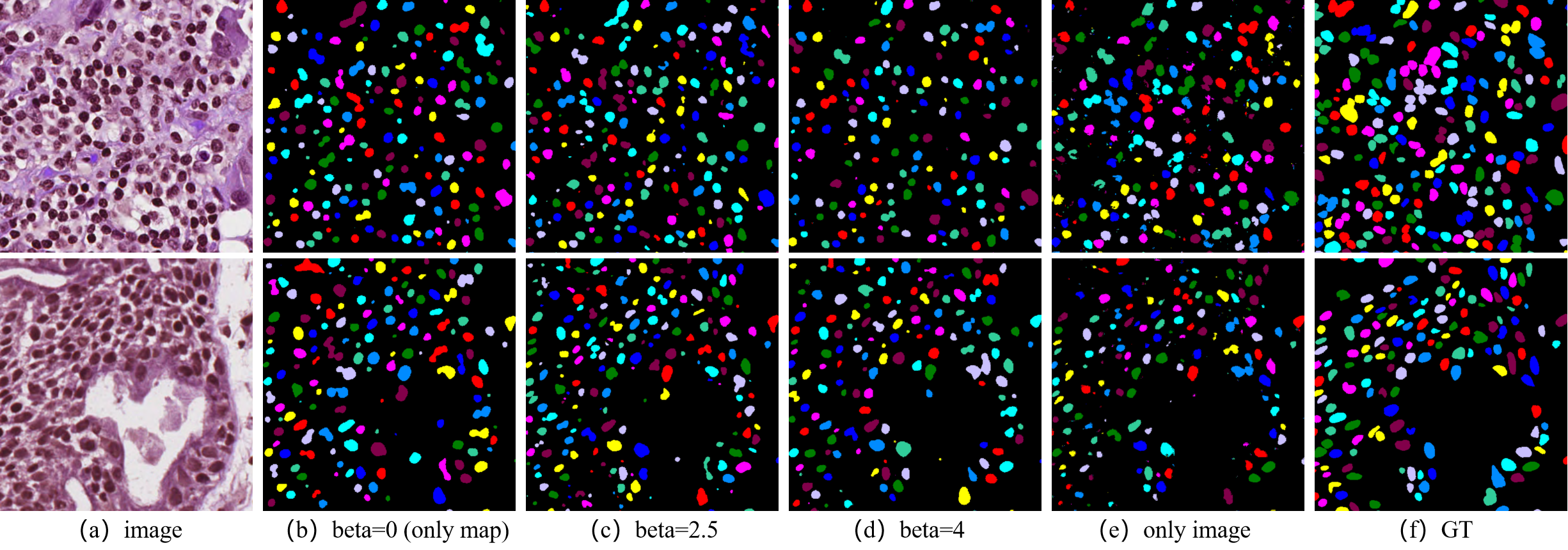}
\caption{Typical segmentation results on MoNuSeg dataset with different detail enhancement ratio $\beta$. (a) image, (b) the results of using the self-activation map itself, (c)-(d) $\beta$ equals to 2.5 or 4, (e) the results of using only the image, (f) ground truth. } \label{fig:beta} 
\end{figure*}

\subsection{Layer Selection} 
Denote the
network with self-supervised training as a cascade of multiple layers
 \begin{equation}
 U_{ss}=(Layer_1,...,Layer_T),
 \end{equation}
 where $T$ is the total number of layers. Note that,
 the definition of each layer is flexible. To be specific, each layer is exactly a block that consists of multiple basic units including ReLU, BacthNorm, and Pooling.
 Every block contains convolutional layers to exploit gradient information and generate self-activation maps.

In our experiments, the backbone can be divided into four layers. We visualize the self-activation map generated from different depths in the model after training with proxy tasks.
We find 
that the self-activation map of shallow layers is sensitive to instance-level properties, as shown in Fig. \ref{fig:layers}. 
The shallow layers contain rich information which is more related to the low-level image features than to the proxy task. 

As demonstrated in Fig. \ref{fig:layers}, high-level semantic representations from the deeper layer reveal the meaningless expression which can hardly 
be used for nuclei detection or segmentation. 
Conversely, low-level semantic representations can provide reliable
descriptions of target cells for training in the segmentation task by highlighting the clear boundaries in the self-activation map.
Moreover, the quantitative evaluation of Fig. \ref{fig:layers_proxy} also reflects the best layer selection is the first block in the backbone.

\subsection{Detail Enhancement} 
The compared experiments demonstrate that adding the information of the raw image can promote the model producing better pseudo masks.
  In Eq.(\ref{detail}), we propose to mix the attention map with the weighted raw information. Hence, adjust the degree of mixing the raw image by controlling the weights, which should be validated in different datasets as shown in Fig. \ref{fig:beta_metric}.
Without the loss of generality, we define a detail enhancement ratio as ${\beta}$. A higher ${\beta}$ value means SGG uses more raw information. When ${\beta}$ is large enough, the mixed image is approximately equal to the input image. 
If $\beta$ is set to zero, it means the self-activation map itself is used for clustering. 
With an increase in the detail enhancement ratio, the metric values drop dramatically after the peak point. The visualization result is demonstrated in Fig. \ref{fig:beta}.



\subsection{Joint Training} 
The training strategy of downstream tasks is also a well-designed part to realize the unsupervised dense object recognition. 
To analyze the effectiveness of joint training, we experiment in the segmentation of MoNuSeg dataset with various training settings. 
As shown in Table \ref{tab:lamda}, the 'w/o' is a baseline implemented by directly training only with the pixel-level label.
The others are implemented by  
adding the object-level supervision from the Voronoi map with different $\lambda$.
Compared with the baseline, a significant improvement is achieved obviously by the joint training mechanism.
Through supervision from pixel-level and object-level pseudo masks, the model can better discriminate the target objects.
The results also reveal that the joint training model is not sensitive to the proportion of object-level supervision. 
When $\lambda$ ranges from 0.5 to 10, the evaluation metrics change very little. 
\begin{table}[!t]
  \centering
  \caption{The effectiveness of joint training module. 'w/o' denotes that we freeze the object-level supervision when training. $\lambda$ indicates the partition enhancement ratio.}
  \scriptsize\begin{tabular}{l | cc | cc}
  \multirow{2}{*}{Method} & \multicolumn{2}{c|}{Pixel-level} & \multicolumn{2}{c}{Object-level} \\
  \cline{2-3} \cline{4-5}
  ~ & IoU & F1 score & Dice & AJI \\
  \hline
    w/o  & 0.501 & 0.722 & 0.623 & 0.390 \\
    \hline
    w/ and ${\lambda}$ = 0.5 &  0.610 & 0.750 & 0.724 & 0.532 \\
    w/ and ${\lambda}$ = 1 & 0.587 & 0.739 & 0.706 & 0.512 \\
    w/ and ${\lambda}$ = 2 & 0.587 & 0.747 & 0.708 & 0.507 \\
    w/ and ${\lambda}$ = 5 & 0.581 & 0.740 & 0.702 & 0.498 \\
    w/ and ${\lambda}$ = 10 & 0.572 & 0.720 & 0.693 & 0.490 \\
  \hline
  \end{tabular}
\label{tab:lamda}
\end{table}

\section{Extension}
\begin{figure*}
\centering
\includegraphics[width=\linewidth]{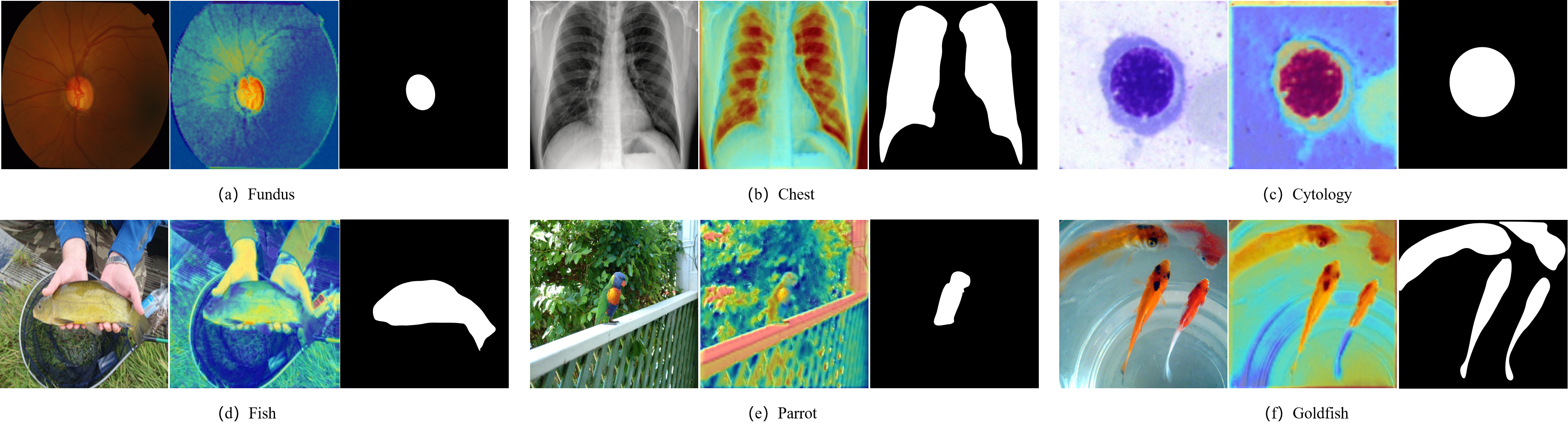}
\caption{Examples of self-activation maps for various types of images. The first row shows the results of different medical images and the bottom row presents natural images. For each block, the first is the original image. The second image is the output of the self-activation map. And the last one is the ground truth.}
\label{fig:others}
\end{figure*}

In this section, the effectiveness of PSM is further verified in various datasets including natural images and other types of biomedical images, which reveals the potential of our method.
Following the procedure in the experiment of nuclei segmentation, we train a self-supervised model using similarity measurement and generate the self-activation map. Results are displayed in Fig. \ref{fig:others}.

\subsection{Image types} 
1) Fundus images: The fundus of the eye includes the optic disc (OD), optic cup (OC).
Our results present a relatively clear boundary in the self-activation map for fundus, which can promote the implementation of unsupervised OD/OC segmentation.

2) Chest X-rays: Chest x-ray reflects internal status of human body, which helps to discover the intrathoracic lesions.
Optimistically, the self-activation map of this image reveals the attention on the inside of the chest.

3) Cytology images:
The segmentation of a single cell is useful for cell recognition and classification which is the key to some disease diagnoses. 
After pre-training the model, clear boundaries of nuclear and the cytoplasm are highlighted in the self-activation map.

4) Natural images. Some types of natural images in the ImageNet are selected. In contrast to medical images, the background in the natural image often contains rich information. Therefore, the attention on the image is not stable in our model. For example, in the picture, a fish is held in the hand and both of them are highlighted. 
Apparently, it is difficult for PSM to focus on  a specific item in a complex scenario without the aid of label information.

\subsection{Weakness and future work}
The quality of generated self-activation maps will decrease 
when the image contains enormous redundant information. 
Moreover, according to the visualization result on natural images, PSM can not cope with the complex scenario containing several categories of targets because the attention may not be allocated at the expected instance.

As demonstrated in Fig. \ref{fig:others}, the PSM is still able to provide relatively suitable self-activation maps for certain tasks such as
OC/OD segmentation of fundus images. 
In the future, we will extend our method to these datasets by looking for suitable supervised learning strategies and introducing new types of pseudo masks.

\section{Conclusion}

In this paper, we proposed a self-activation map based framework for unsupervised nuclei detection and segmentation. The framework is composed of a prior self-activation module (PSM), a semantic-guided generator 
(SGG), a nuclei detection network (NDN) and a nuclei segmentation network (NSN). 
The proposed PSM has a strong capability of learning low-level representations to highlight the area of interest without the need for manual labels. We also designed SGG that serves as a pipeline between representation features from the self-supervised network and the downstream segmentation task. And NDN and NSN can strengthen the model's ability to identify intensive objects and achieve pixel-level segmentation. Our unsupervised method was evaluated on two publicly available pathological datasets and obtained competitive results compared to the methods with annotations. The results also showed that our framework had the potential in processing other types of images.








\end{document}